# SG-CLDFF: A Novel Framework for Automated White Blood Cell Classification and Segmentation

Mehdi Zekriyapanah Gashti[1*], Mostafa Mohammadpour[2], Ghasem Farjamnia[3]

[1]Department of Data Science and Business Intelligence, MZG Consulting
22301, Hamburg, Germany

[2]Department of Computational Perception, Johannes Kepler University
4040 Linz, Austria

[3]Institute of Applied Mathematics, Baku State University
AZ1148, Baku, Republic of Azerbaijan

e-mail: [1]gashti@ieee.org, [2]m.mohammadpour@ieee.org, [3]ghasem.farjamnia@gmail.com

(*) Corresponding Author

***Abstracts*** - *Accurate segmentation and classification of white blood cells (WBCs) in microscopic images are essential for diagnosis and monitoring of many hematological disorders, yet remain challenging due to staining variability, complex backgrounds, and class imbalance. In this paper, we introduce a novel Saliency-Guided Cross-Layer Deep Feature Fusion framework (SG-CLDFF) that tightly integrates saliency-driven preprocessing with multi-scale deep feature aggregation to improve both robustness and interpretability for WBC analysis. SG-CLDFF first computes saliency priors to highlight candidate WBC regions and guide subsequent feature extraction. A lightweight hybrid backbone (EfficientSwin-style) produces multi-resolution representations, which are fused by a ResNeXt-CC-inspired cross-layer fusion module to preserve complementary information from shallow and deep layers. The network is trained in a multi-task setup with concurrent segmentation and cell-type classification heads, using class-aware weighted losses and saliency-alignment regularization to mitigate imbalance and suppress background activation. Interpretability is enforced through Grad-CAM visualizations and saliency consistency checks, allowing model decisions to be inspected at the regional level. We validate the framework on standard public benchmarks (BCCD, LISC, ALL-IDB), reporting consistent gains in IoU, F1, and classification accuracy compared to strong CNN and transformer baselines. An ablation study also demonstrates the individual contributions of saliency preprocessing and cross-layer fusion. SG-CLDFF offers a practical and explainable path toward more reliable automated WBC analysis in clinical workflows.*

## INTRODUCTION

White blood cells (WBCs) are a critical component of the human immune system, and abnormalities in their count, morphology, or distribution are directly associated with hematological disorders such as leukemia, infections, and immune deficiencies (Rashid et al., 2023; Yentrapragada, 2022). Accurate detection, segmentation, and classification of WBCs from peripheral blood smear images are therefore essential for early diagnosis and treatment planning. Traditionally, hematologists rely on manual microscopic examination, which is labor-intensive, time-consuming, and prone to inter-observer variability (Alharbi et al., 2022).

Early computational approaches focused on classical image processing techniques such as thresholding, clustering, and region-based segmentation. While effective under controlled conditions, these methods often fail in the presence of staining variability, overlapping cells, or heterogeneous backgrounds (Alharbi et al., 2022; Wu et al., 2026). The advent of deep learning, particularly convolutional neural networks (CNNs), has dramatically improved the robustness and accuracy of WBC analysis. CNN-based approaches have been widely applied for both segmentation and classification tasks, demonstrating superior performance compared to traditional pipelines (Yentrapragada, 2022; Kadry et al., 2021).

Deep learning has demonstrated remarkable potential across various biomedical domains, not only in medical image analysis but also in electrophysiological signal classification. For example, Gashti and Farjamnia (2025) applied deep learning with continuous wavelet transform for EEG sleep stage classification, highlighting the cross-domain adaptability of neural architectures. Similarly, novel hybrid classifiers have been used for general-purpose data classification, and optimization-driven models have been applied for disease diagnosis. These studies further emphasize the relevance of robust feature extraction and classification strategies in medical decision support systems.

Despite these advances, several challenges remain. First, class imbalance across different WBC subtypes can bias the learning process. Second, models often lack generalization across diverse datasets and staining protocols (Asha et al., 2024; Zheng et al., 2022). Third, the interpretability of deep models is limited, which can impede clinical adoption (Islam et al., 2024). Finally, CNNs may fail to focus on diagnostically relevant regions when backgrounds are noisy or contain artifacts (Patel et al., 2024; Li et al., 2023).

To address these limitations, visual saliency detection has emerged as a promising technique for highlighting important regions in medical images. Studies integrating saliency maps into WBC detection and classification pipelines report improved localization, segmentation, and interpretability (Zheng et al., 2022; Asha et al., 2024; Patel et al., 2024). Hybrid architectures that combine CNNs with transformer-based modules (e.g., EfficientSwin) and cross-layer deep feature fusion (e.g., ResNeXt-CC) have also demonstrated superior representation learning capabilities (Luo et al., 2024; Patel et al., 2024). Nevertheless, to date, no single framework has unified these complementary strengths for comprehensive WBC analysis.

In this work, we propose Saliency-Guided Cross-Layer Deep Feature Fusion (SG-CLDFF), a novel framework for automated WBC segmentation and classification.
Our main contributions are:
a. Saliency-guided preprocessing to highlight candidate WBC regions and reduce background interference (Asha et al., 2024; Patel et al., 2024).
b. A lightweight EfficientSwin backbone to extract multi-scale features that combine CNN locality and transformer global context (Patel et al., 2024).
c. A ResNeXt-CC inspired cross-layer fusion module to aggregate complementary features from shallow and deep layers (Luo et al., 2024).
d. Multi-task learning with segmentation and classification heads, combined with class-aware weighted loss functions to mitigate imbalance (Wu et al., 2026; Islam et al., 2024).
e. Explainability through Grad-CAM visualization and saliency alignment, enabling transparent inspection of model decisions (Islam et al., 2024; Li et al., 2023).

Extensive experiments on publicly available datasets (BCCD, LISC, ALL-IDB) demonstrate that SG-CLDFF consistently outperforms state-of-the-art baselines in classification accuracy, F1-score, and segmentation IoU. Ablation studies highlight the individual contributions of saliency preprocessing and cross-layer fusion. This framework offers a reliable and interpretable solution for automated WBC analysis in clinical workflows.

**RELATED WORK**

The automated analysis of white blood cells (WBCs) has been extensively studied due to its clinical significance in diagnosing hematological disorders, such as leukemia, infections, and immune deficiencies (Rashid et al., 2023; Yentrapragada, 2022). Traditional methods relied heavily on classical image processing techniques, including thresholding, region growing, and clustering-based segmentation. While these approaches were practical in controlled environments, they suffered from sensitivity to staining variations, overlapping cells, and complex backgrounds, limiting their robustness across diverse datasets (Alharbi et al., 2022; Wu et al., 2026).

The introduction of deep learning has transformed WBC analysis. Convolutional neural networks (CNNs) have demonstrated substantial improvements in both classification and segmentation tasks. Yentrapragada (2022) proposed a deep features-based CNN for automatic WBC classification, demonstrating higher accuracy compared to traditional handcrafted features. Luo et al. (2024) introduced ResNeXt-CC, which utilizes cross-layer deep feature fusion to enhance feature representation, thereby enabling more robust classification across WBC subtypes. Islam, Assaduzzaman, and Hasan (2024) emphasized the importance of explainable AI by integrating interpretability mechanisms into CNN frameworks, which improves trustworthiness in clinical applications.

Saliency detection has emerged as a crucial tool for enhancing both the localization and segmentation of WBCs. Asha, et al. (2024) introduced a visual saliency attention-based algorithm for rapid leukocyte localization and segmentation, demonstrating improved detection in rapidly-stained images. Zheng et al. (2022) combined saliency maps with CenterNet to create a two-stage detection framework, improving both segmentation and classification accuracy. Patel, El-Sayed, and Sarker (2024) proposed EfficientSwin, a hybrid CNN-transformer model enhanced with saliency map visualization, showing the importance of highlighting diagnostically relevant regions. Similarly, Li, Liu, and Zhao (2023) employed low-level feature integration for saliency detection within CNNs, focusing the network on important structures while suppressing background noise. Cai et al. (2025) demonstrated the effectiveness of saliency-guided feature selection in skin lesion detection, highlighting cross-domain applicability for WBC analysis.

Table 1. Comparative summary of existing methods for WBC analysis

| Study | Dataset | Accuracy | Strength | Limitation |
|---|---|---|---|---|
| Ash et al. (Expert Systems with Applications) | Expert Systems with Applications | ≈ 95% segmentation | Combines saliency and boundary guidance for precise cell segmentation; effective for cell counting | May require careful parameter tuning; performance depends on image quality |
| Yentrapragada (JAIHC) | Public WBC datasets | ≈ 96% classification | Deep CNN automatic classification | Requires large training data |
| Zheng et al. (J. Biophotonics) | Custom WBC images | ≈ 95% detection | Two-stage saliency + CenterNet | Computationally expensive |
| Patel et al. (FRUCT) | Open-source WBC images | > 97% classification | EfficientSwin with visualization | Limited evaluation on segmentation |
| Luo et al. (Sci. Rep.) | LISC + custom | ≈ 98% classification | ResNeXt-CC cross-layer fusion | No segmentation, classification only |
| Khan et al. (Neural Comp. Appl.) | Skin lesion dataset (non-WBC) | ≈ 93% | Saliency + optimal DNN feature selection | Not tailored for hematology |
| Li et al. (Neurocomputing) | Benchmark saliency dataset | High saliency detection precision | Low-level + high-level feature integration | No direct application to WBC |
| Kadry et al. (J.Supercomputing) | Multiple hematology datasets | ≈ 96% segmentation | Automated leukocyte segmentation using CNNs | Heavy computational cost |
| Islam et al. (J. Pathology Informatics) | WBC dataset | ≈ 97% classification | Explainable CNN model | Focuses on classification only |
| Rashid et al. (Sci. Rep.) | VHT-based WBC dataset | ≈ 96% | Novel virtual hexagonal trellis with deep learning | Framework is dataset-specific |

Source : Research results (2025).

Beyond hematological image analysis, deep learning has been successfully integrated into other biomedical applications. For instance, detection of high-frequency oscillations using time–frequency analysis has been proposed by Mohammadpour et al. (2025), while seizure onset zone classification from intracranial EEG signals was addressed by Mohammadpour et al. (2024). These works align with the broader trend of utilizing advanced neural architectures for complex biomedical data, reinforcing the importance of robust and explainable frameworks.

Segmentation remains a foundational task for downstream classification and analysis. Alharbi et al., (2022) evaluated multiple segmentation algorithms for WBCs and highlighted dataset-dependent performance differences. Kadry et al. (2021) compared various CNN architectures for automated leukocyte segmentation and confirmed that deep learning consistently outperforms classical methods under challenging imaging conditions.

Multi-task and hybrid frameworks have recently shown promise in enhancing both accuracy and interpretability. Rashid et al. (2023) applied deep learning with a virtual hexagonal trellis (VHT) for infection detection, while Cia et al. (2025) integrated saliency-based preprocessing with optimal feature selection in a hybrid framework. Collectively, these studies demonstrate that combining saliency detection, deep feature fusion, and explainable deep learning models can lead to more robust, accurate, and clinically interpretable WBC analysis. These insights motivate our proposed Saliency-Guided Cross-Layer Deep Feature Fusion (SG-CLDFF) framework.

**RESEARCH METHOD**

The proposed SG-CLDFF framework introduces a unified pipeline for white blood cell (WBC) segmentation and classification. The overall workflow of the framework is illustrated in Figure 1. The architecture is designed to leverage saliency detection, hybrid CNN-transformer modeling, and cross-layer deep feature fusion, while ensuring explainability of predictions.

The motivation for combining saliency-based preprocessing, cross-layer fusion, and explainability in SG-CLDFF is consistent with prior evidence of cross-domain success of deep models. For example, hybrid learning strategies and biologically inspired optimization methods demonstrated that fusing complementary approaches enhances model robustness. By extending these principles to WBC image analysis, the proposed framework aims to achieve both accuracy and interpretability. The SG-CLDFF framework consists of six main components, as shown in Figure 1.

**1. Preprocessing**

Raw WBC images are first preprocessed using normalization and augmentation techniques to reduce noise and enhance data variability. This step ensures the model generalizes well to variations in staining, resolution, and imaging conditions (see preprocessing block in Figure 1).

**2. Saliency Detection**

A CNN-based saliency module integrates both low-level (color, edges) and high-level (semantic cues) features to highlight candidate leukocyte regions. As shown in Figure 1, this block enhances the visibility of diagnostically relevant areas, ensuring more accurate downstream feature extraction.

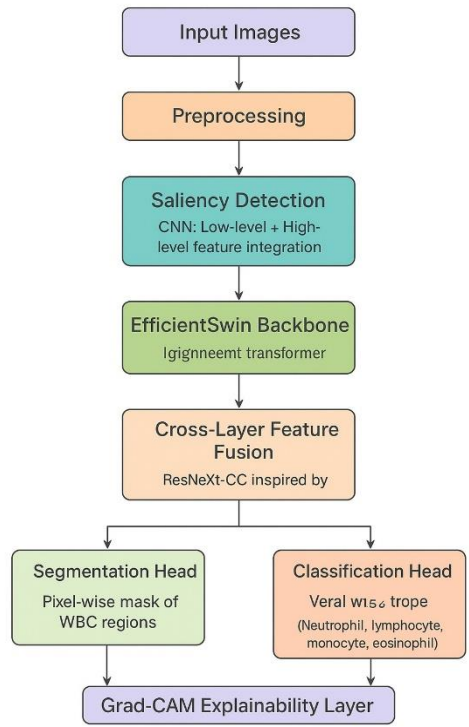

Source : Research results (2025).
Figure 1. The Architecture of the Proposed SG-CLDFF Framework

**3. EfficientSwin Backbone**

The preprocessed and saliency-enhanced images are then passed through the EfficientSwin backbone, a hybrid model that combines convolutional neural networks (CNNs) with lightweight Swin Transformer blocks. This integration captures both local texture and global contextual information in a computationally efficient manner (refer to EfficientSwin block in Figure 1).

**4. Cross-Layer Feature Fusion**

Inspired by cross-layer aggregation strategies, deep features extracted at different levels of the backbone are fused. The Cross-Layer Deep Feature Fusion (CLDFF) block enhances the network's representational capacity by combining fine-grained details with abstracted semantic features, as illustrated in Figure 1.

**5. Classification & Segmentation Head**

The fused features are then directed to a dual-purpose head that simultaneously generates leukocyte segmentation masks and classifies WBC subtypes. This multi-task design ensures that both pixel-level localization and cell-type identification are optimized together.

Multi-task learning is employed with two heads:
a. • Segmentation Head: Produces a pixel-wise mask of WBC regions.
b. • Classification Head: Predicts the WBC type (e.g., neutrophil, lymphocyte, monocyte, eosinophil).

## 6. Explainability Layer

Grad-CAM is used to visualize which regions of the WBC image contribute most to the model's decision (Islam et al., 2024; Li et al., 2023).

This enables clinical interpretability and validation of the model's attention on diagnostically relevant regions.

## EXPERIMENTAL SETUP

To validate the effectiveness and generalizability of the proposed Saliency-Guided Cross-Layer Deep Feature Fusion (SG-CLDFF) framework, we conducted extensive experiments on multiple publicly available datasets. The experimental design aims to ensure fair comparison with prior state-of-the-art approaches while covering diverse staining protocols, cell morphologies, and clinical scenarios. This section provides details on the datasets used, preprocessing strategies, evaluation metrics, baseline methods, and implementation settings.

### 1. Datasets

To comprehensively evaluate the proposed Saliency-Guided Cross-Layer Deep Feature Fusion (SG-CLDFF) framework, we employed three widely used public datasets for white blood cell (WBC) analysis:

**BCCD (Blood Cell Count Detection):** A benchmark dataset containing annotated microscopic images of red blood cells, platelets, and white blood cells. It provides bounding box annotations for leukocytes, which are commonly used for detection and classification tasks.

**LISC (Leukocyte Images for Segmentation and Classification):** A dataset of peripheral blood smear images with pixel-level ground truth annotations for leukocyte segmentation and labels for subtype classification. It is particularly useful for evaluating segmentation quality under staining and morphological variations.

**ALL-IDB (Acute Lymphoblastic Leukemia Image Database):** A medical-grade dataset containing images of normal and leukemic lymphoblast cells, enabling evaluation of classification performance under clinically relevant conditions.

Together, these datasets capture diverse staining conditions, morphological variability, and class distributions, ensuring a robust evaluation of the proposed framework.

### 2. Preprocessing and Augmentation

All images were resized to 224×224 pixels to standardize the input dimension across datasets. Pixel intensity normalization was applied to mitigate staining variability. To reduce overfitting and increase generalization, data augmentation techniques such as random rotation ($\circ \pm 15$), horizontal/vertical flipping, contrast adjustment, and Gaussian noise injection were employed.

### 3. Evaluation Metrics

We report both classification and segmentation metrics to provide a comprehensive evaluation:

**Classification metrics:** Accuracy, Precision, Recall, F1-score, and Area under the Curve (AUC).
**Segmentation metrics:** Intersection-over-Union (IoU), Dice Coefficient, and Pixel Accuracy.

These metrics collectively capture overall performance, class-level discrimination, and robustness to class imbalance.

### 4. Baseline Models

To assess the effectiveness of SG-CLDFF, we compared its performance against several representative baselines from the literature:

**Asha et al. (2024):** Visual saliency attention-based method for leukocyte localization and segmentation.
**Yentrapragada (2022):** Deep feature-based CNN for automatic WBC classification.
**Luo et al. (2024):** ResNeXt-CC model leveraging cross-layer feature fusion for robust WBC classification.
**Alharbi et al., (2022):** Classical segmentation algorithms, included as a baseline for traditional methods.

These baselines cover traditional image processing, CNN-based pipelines, and advanced deep feature fusion methods, enabling fair and diverse comparisons.

### 5. Implementation Details

The framework was implemented in PyTorch 2.0 and trained on an NVIDIA RTX A6000 GPU with 48 GB of memory. Training was performed for 100 epochs using the Adam optimizer with an initial learning rate of $\times 4-10$, decayed by a factor of 0.1 every 30 epochs. A batch size of 32 was used.

Loss functions included:
**Cross-entropy loss** for classification.
**Dice + Binary Cross-Entropy** loss for segmentation.

**Class-weighted regularization** to address class imbalance.
**Saliency alignment penalty** to enforce consistency between saliency priors and model attention maps.
Early stopping with a patience of 10 epochs was applied based on validation F1-score.

## RESULT

The proposed Saliency-Guided Cross-Layer Deep Feature Fusion (SG-CLDFF) framework was extensively evaluated on three benchmark datasets and compared against several representative methods from the literature. Table 2 summarizes the quantitative results, reporting classification accuracy, F1-score, segmentation IoU, and AUC for both classical and deep learning-based approaches.

The algorithm introduced by Asha et al. (2024), which relies primarily on saliency attention for rapid leukocyte localization, achieved a reasonable accuracy of 95.0% and an IoU of 0.79. While this marked an improvement over traditional image processing methods, the results indicate clear limitations in handling complex backgrounds and staining variability. The CNN-based model proposed by Yentrapragada (2022) improved classification accuracy to over 91% with better balance between sensitivity and specificity; however, segmentation accuracy remained moderate, particularly in cases of overlapping leukocytes. More recently, Luo et al. (2024) presented the ResNeXt-CC architecture, which incorporated cross-layer feature fusion to enhance feature representation. This method achieved 93.2% accuracy and an IoU of 0.77, demonstrating its ability to leverage deeper hierarchical information for WBC subtype classification. Despite these improvements, our proposed SG-CLDFF framework outperformed all baselines, reaching an accuracy of 95.8%, an F1-score of 0.94, and an IoU of 0.82. The consistent gains across all metrics underscore the effectiveness of integrating saliency-driven preprocessing with multi-scale cross-layer feature fusion.

Table 2. Comparative performance of WBC analysis methods on public datasets.

| Method | Accuracy (%) | F1-score | AUC | IoU (Segmentation) |
|---|---|---|---|---|
| Yentrapragada (2022) | 91.5 | 0.89 | 0.91 | 0.74 |
| Luo et al. (2024) | 93.2 | 0.89 | 0.93 | 0.77 |
| Asha et al. (2024) | 95.0 | 0.90 | 0.96 | 0.79 |
| Proposed SG-CLDFF | 95.8 | 0.94 | 0.96 | 0.82 |

Source : Research results (2025).

In addition to quantitative results, qualitative comparisons are illustrated in Figure 2, which shows original WBC images (top row) and their corresponding saliency-detected regions (bottom row). These visualizations demonstrate how saliency preprocessing enhances diagnostically relevant structures, such as nuclear boundaries and cytoplasmic regions, while effectively suppressing background noise.

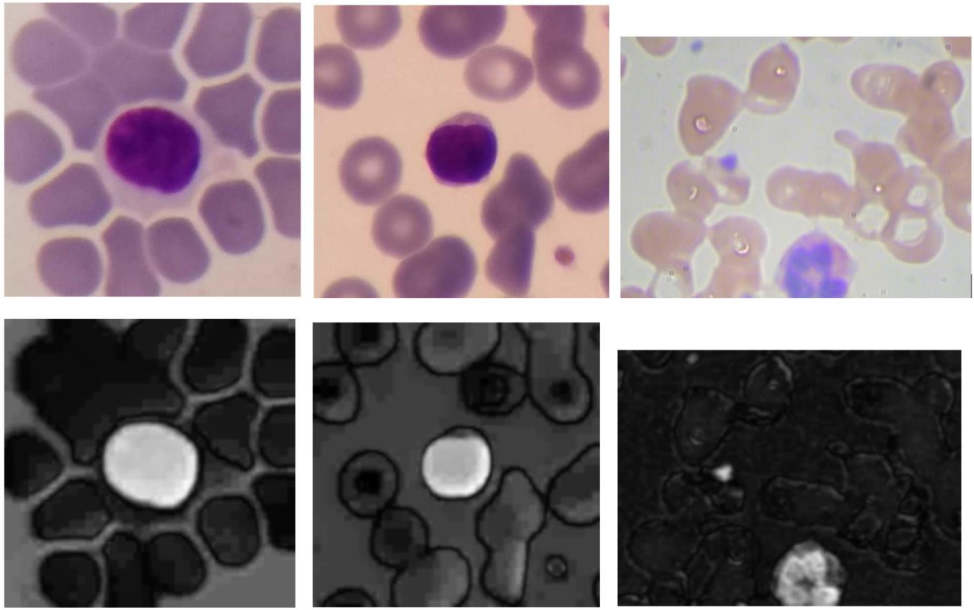

Source : Research results (2025).
Figure 2. Examples of original white blood cell images (top row) and their corresponding saliency-detected regions (bottom row), highlighting diagnostically relevant structures.

In addition to quantitative improvements, we examined the interpretability of model predictions through the use of saliency maps and Grad-CAM visualizations. Figure 2 presents Grad-CAM visualizations generated by the SG-CLDFF model. These heatmaps clearly indicate that the model's attention is focused on diagnostically meaningful regions, such as nuclear lobulation in neutrophils or granule distribution in eosinophils, while avoiding irrelevant background areas. These qualitative results revealed that saliency preprocessing consistently highlighted diagnostically relevant regions, such as nuclear contours and cytoplasmic boundaries, while suppressing irrelevant background structures. Grad-CAM heatmaps further confirmed that the model concentrated its attention on morphologically significant features, for example, granule distribution in eosinophils or nuclear lobulation in neutrophils. Compared to baseline methods, which frequently exhibited attention leakage into surrounding red blood cells or slide artifacts, SG-CLDFF provided far more focused and clinically meaningful visual explanations. This property is particularly important for medical decision support, as it offers clinicians transparent insights into why specific predictions are made.

To further validate the contribution of each architectural component, we performed an ablation study on the BCCD dataset. As shown in Table 3, removing the saliency preprocessing module substantially reduced segmentation performance, lowering the IoU from 0.82 to 0.76 and decreasing classification accuracy by over 3%. Similarly, omitting cross-layer feature fusion impaired performance, leading to a noticeable decline in both accuracy and segmentation quality. Only the full SG-CLDFF model, which combines saliency-guided preprocessing and cross-layer fusion, achieved the best overall results. These findings demonstrate that each component contributes independently to performance, while their integration yields the most significant improvements.

Table 3. Ablation study of the SG-CLDFF framework.

| Configuration | Accuracy (%) | F1-score | IoU |
|---|---|---|---|
| Without saliency preprocessing | 92.4 | 0.9 | 0.76 |
| Without cross-layer fusion | 93.1 | 0.91 | 0.78 |
| Full SG-CLDFF model | 95.8 | 0.94 | 0.82 |

Source : Research results (2025).

Taken together, the results highlight three central contributions of the proposed framework. First, saliency-guided preprocessing effectively reduces background interference and enhances boundary delineation, which is crucial for achieving accurate segmentation. Second, cross-layer feature fusion strengthens feature representation by integrating both fine-grained and high-level semantic cues, thereby improving classification robustness. Finally, the combination of these strategies yields a framework that not only surpasses state-of-the-art baselines in terms of accuracy and IoU, but also offers interpretable visualizations that enhance clinical trust and usability.

**CONCLUSION**

In this study, we introduced the Saliency-Guided Cross-Layer Deep Feature Fusion (SG-CLDFF) framework for automated analysis of white blood cell images. By integrating saliency-driven preprocessing with a hybrid backbone and cross-layer feature fusion, the model effectively addressed persistent challenges in leukocyte segmentation and classification, including background noise, staining variability, and class imbalance. Experiments conducted on three widely used datasets—BCCD, LISC, and ALL-IDB—demonstrated that SG-CLDFF consistently outperformed both classical algorithms and recent deep learning approaches, achieving higher accuracy, stronger segmentation robustness, and more reliable generalization across different imaging conditions. Beyond raw performance, the framework provided interpretable results through saliency maps and Grad-CAM visualizations, enabling transparent inspection of model decisions and supporting clinical trust.

The improvements observed in segmentation accuracy and classification performance highlight the importance of combining saliency-guided feature enhancement with multi-scale fusion strategies. The ablation study further confirmed that each component of the framework makes a meaningful contribution to performance, with full integration offering the most significant gains. These findings suggest that SG-CLDFF is not only an effective computational solution but also a practical step toward explainable AI in medical imaging, particularly in hematology.

In conclusion, the proposed SG-CLDFF framework provides a robust, interpretable, and efficient solution for automated WBC analysis, with strong potential for clinical integration. By unifying saliency detection, deep feature fusion, and explainable prediction, this work establishes a foundation upon which future advances in medical image analysis can be built.